\documentclass[10pt,twocolumn,letterpaper]{article}

\usepackage[pagenumbers]{cvpr}

\ifdefined\pdfminorversion
\fi
\ifdefined\pdfobjcompresslevel
\fi

\usepackage{amsmath,amssymb,amsfonts}
\usepackage{booktabs}
\usepackage{multirow}
\usepackage{xspace}
\usepackage{subcaption}
\usepackage{algorithm}
\usepackage{algorithmic}

\newcommand{\method}{Le MuMo JEPA\xspace}

\definecolor{cvprblue}{rgb}{0.21,0.49,0.74}
\usepackage[pagebackref,breaklinks,colorlinks,allcolors=cvprblue]{hyperref}

\def\paperID{29}
\def\confName{CVPR}
\def\confYear{2026}

\title{Le MuMo JEPA: Multi-Modal Self-Supervised Representation Learning\\with Learnable Fusion Tokens}
\author{Ciem Cornelissen\qquad Sam Leroux\qquad Pieter Simoens\\
IDLab, Department of Information Technology, Ghent University - imec\\
Belgium\\
{\tt\small \{ciem.cornelissen, sam.leroux, pieter.simoens\}@ugent.be}}

\begin{document}
\maketitle

\begin{abstract}
Self-supervised learning has emerged as a powerful paradigm for learning visual representations without manual annotations, yet most methods still operate on a single modality and therefore miss the complementary structure available from heterogeneous sensors.
We present \method, a self-supervised framework that learns unified representations from RGB images and aligned companion modalities.
In our driving experiments, the second modality is camera-aligned LiDAR depth; we also evaluate RGB-thermal training and transfer on the Teledyne FLIR ADAS benchmark.
Our approach extends LeJEPA to the multi-modal setting by learning \emph{fusion tokens} that act as a latent bottleneck between modality-specific patch stems inside a shared transformer.
Our default model employs a pruned fusion strategy: after an initial cross-modal attention layer, modality-specific tokens are dropped, forcing cross-modal information into the shared fusion-token grid as an efficient latent bottleneck before Sketched Isotropic Gaussian Regularization (SIGReg) is applied to the joint multimodal CLS embedding.
On Waymo, \method gives the strongest performance-efficiency trade-off on downstream patch probes among the from-scratch multimodal baselines, improving CenterNet detection and dense depth while remaining competitive on segmentation.
Under from-scratch training on nuScenes, \method remains the strongest model, and it also gives the best FLIR results, especially after Waymo-initialized fine-tuning.
It also retains the best overall accuracy-efficiency balance in our study at substantially lower compute, memory, and estimated training time.
\end{abstract}
\section{Introduction}
\label{sec:intro}

Many real-world perception systems rely on multiple sensors, with cameras providing dense texture and color information while complementary sensor modalities such as LiDAR depth or thermal infrared contribute geometry, temperature, or range cues.
Learning representations that effectively combine these signals remains an open challenge, and most state-of-the-art multi-modal perception models~\cite{liu2023bevfusion, bai2022transfusion} are still trained in a fully supervised manner with costly large-scale annotations.
Autonomous driving provides a natural testbed for this paradigm because modern perception stacks already rely on paired dense sensors and spatial correspondence across those streams matters directly for downstream reasoning.

Self-supervised learning (SSL) offers a compelling alternative by learning general-purpose representations from unlabeled data.
Methods such as BYOL~\cite{grill2020byol}, DINO~\cite{caron2021dino}, MAE~\cite{he2022mae}, and I-JEPA~\cite{assran2023ijepa} have achieved strong image-understanding results. LeJEPA~\cite{balestriero2025lejepa} further introduces \emph{Sketched Isotropic Gaussian Regularization} (SIGReg), which constrains embeddings to follow an isotropic Gaussian distribution without relying on common SSL heuristics such as stop-gradients or teacher-student networks. For multi-modal learning, this is appealing because both modalities are pulled toward the same data-agnostic target geometry instead of being aligned only through pairwise contrastive matching, which typically depends more heavily on negative sampling and careful control of the modality gap.

\begin{figure}[t]
    \centering
    \includegraphics[width=\linewidth]{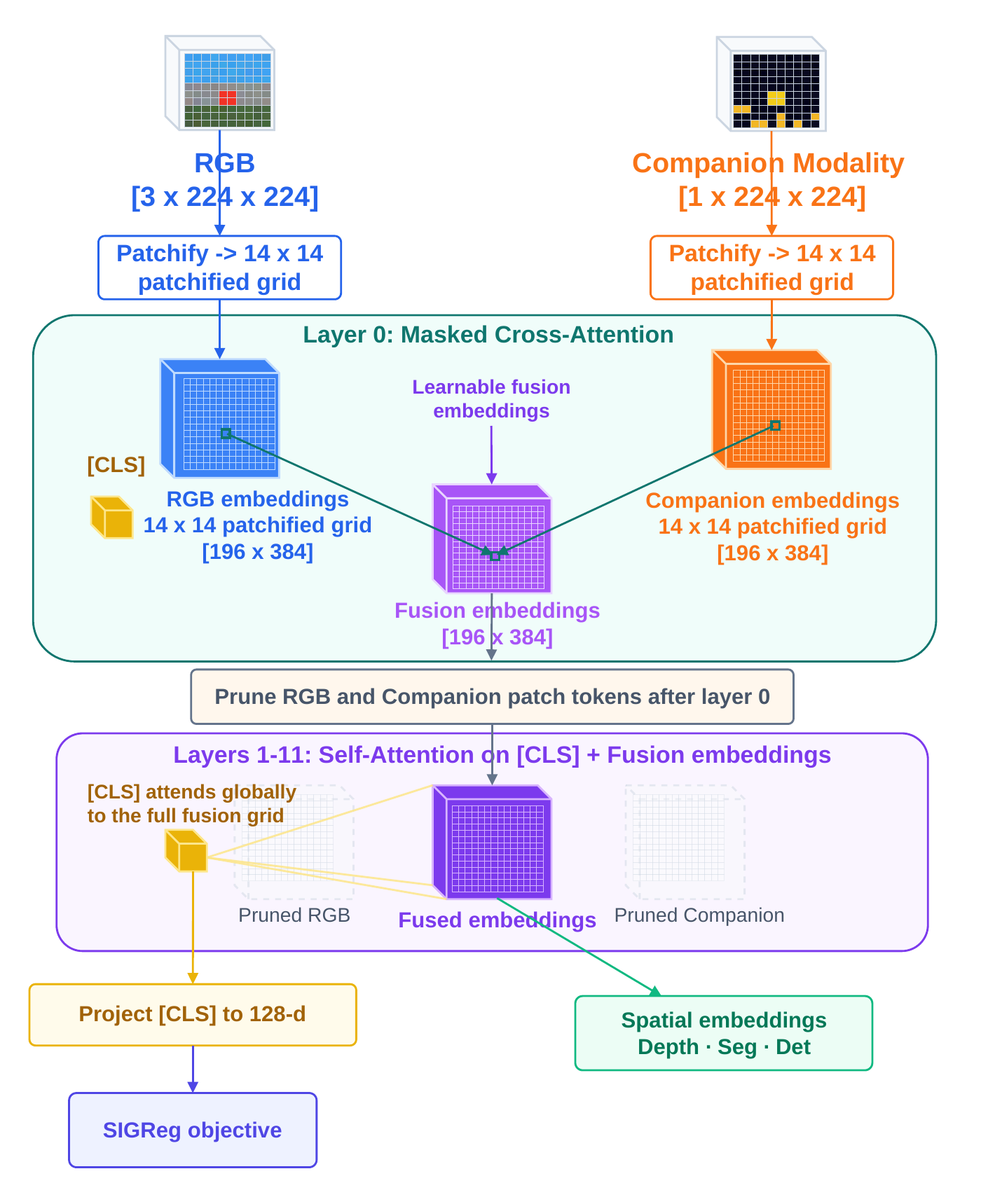}
    \caption{\textbf{Overview of \method.} The companion modality is represented as a spatially aligned signal and fused with RGB through learnable \emph{fusion tokens}, which act as a latent bottleneck inside a shared transformer. The default training objective applies SIGReg to the joint multimodal CLS embedding.}
    \label{fig:architecture}
\end{figure}

However, existing JEPA-based methods operate exclusively on single-modality inputs.
Extending self-supervised learning to RGB-plus-modality settings raises a more specific problem: the streams have very different structures; they must still be aligned well enough to support meaningful correspondences, and their interaction must be explicit and efficient, avoiding both the representational limits of weak late fusion and the quadratic computational cost of unrestricted all-to-all token mixing.
We therefore frame both streams on a shared 2D spatial grid so that dense token-to-token alignment happens inside one transformer without the overhead of a separate sparse 3D backbone~\cite{liu2023bevfusion,yang2024unipad}. Keeping both modalities in one token space also makes it easier to transfer from RGB-LiDAR training to RGB-thermal evaluation on FLIR~\cite{flir_adas_v2}.

In this work, we present \method, a self-supervised framework that learns unified RGB-plus-companion-modality representations.
Our key insight is that SIGReg provides a natural \emph{modality-agnostic} shared target: by encouraging both RGB and companion-modality representations to follow the same Gaussian reference distribution, the model learns a shared embedding geometry across modalities without imposing artificial token-wise pairing constraints.

Within a standard Vision Transformer~\cite{dosovitskiy2021vit}, we introduce \emph{learnable fusion tokens}, a set of tokens equal in number to the spatial patches that aggregate information from spatially corresponding RGB and companion-modality patches through attention.
Conceptually, these tokens form a Perceiver-style latent bottleneck~\cite{jaegle2021perceiverio}: cross-modal communication is routed through a learned spatial memory buffer instead of allowing unrestricted all-to-all interaction between every RGB and companion-modality token.

We evaluate \method on Waymo~\cite{sun2020waymo}, nuScenes~\cite{caesar2020nuscenes}, and RGB-thermal FLIR~\cite{flir_adas_v2}.
Waymo and nuScenes use matched from-scratch training within each dataset, while FLIR is reported both from scratch and under Waymo-initialized transfer and fine-tuning.
Using frozen patch probes for CenterNet-style 3D object detection~\cite{zhou2019centernet}, depth estimation, and segmentation mIoU, we show that the default \method configuration consistently outperforms single-modality baselines and simpler early- or late-fusion alternatives.
Waymo and nuScenes comparisons retrain each encoder from scratch within the target dataset rather than comparing against off-the-shelf checkpoints; FLIR additionally includes Waymo-to-FLIR transfer and fine-tuning.
The largest gains appear in object-centric localization and depth, where explicit cross-modal fusion improves both geometry and downstream patch-probe performance over RGB-only JEPA~\cite{balestriero2025lejepa}, a DINOv3-style RGB baseline~\cite{simeoni2025dinov3}, MultiMAE~\cite{bachmann2022multimae}, ImageBind~\cite{girdhar2023imagebind}, and simple fusion ablations.

Our contributions are:
\begin{itemize}[leftmargin=1.3em]
    \item We extend LeJEPA to the multi-modal setting and show that SIGReg regularization supports joint RGB-plus-modality representation learning without auxiliary alignment labels.
    \item We introduce learnable fusion tokens with a pruned default design that acts as an efficient latent bottleneck, together with a persistent-routing ablation, and show that applying SIGReg to the joint multimodal CLS embedding yields the strongest default configuration in our study.
    \item We benchmark \method on Waymo, nuScenes, and FLIR against single-modality controls, early and late fusion ablations, MultiMAE, a DINOv3-style RGB baseline, and ImageBind under matched from-scratch protocols on Waymo and nuScenes, plus from-scratch and Waymo-initialized evaluation on FLIR, using the same frozen patch probes and compute profiling pipeline.
\end{itemize}

\section{Related Work}
\label{sec:related}

\paragraph{Self-Supervised Visual Representation Learning.}
Self-supervised visual learning spans contrastive methods such as SimCLR~\cite{chen2020simclr} and MoCo~\cite{he2020moco}, non-contrastive methods such as BYOL~\cite{grill2020byol} and VICReg~\cite{bardes2022vicreg}, masked modeling methods such as MAE~\cite{he2022mae} and BEiT~\cite{bao2022beit}, and JEPA-style latent prediction~\cite{assran2023ijepa, bardes2024vjepa}.
LeJEPA~\cite{balestriero2025lejepa} provides theoretical foundations for JEPA models through SIGReg, which matches the embedding distribution to $\mathcal{N}(0, \mathbf{I})$.
Our work extends LeJEPA to multimodal RGB-plus-companion-modality inputs while retaining its regularization framework and JEPA-style predictive training.

\paragraph{Multi-Modal Sensor Fusion.}
Supervised camera-LiDAR fusion has been explored at various stages, from point-level sequential fusion~\cite{vora2020pointpainting} to late-stage bounding-box candidate fusion~\cite{pang2020clocs}, alongside feature-space systems using transformers or bird's-eye view (BEV) grids such as BEVFusion~\cite{liu2023bevfusion, liang2022bevfusion} and TransFusion~\cite{bai2022transfusion}. These methods rely on extensive 3D annotations.
In contrast, we learn fused representations in a fully self-supervised manner, requiring no labeled data during pretraining.

\paragraph{Self-Supervised Multi-Modal Learning.}
Multi-modal self-supervised learning includes shared-embedding alignment methods such as ImageBind~\cite{girdhar2023imagebind}, masked reconstruction methods such as MultiMAE~\cite{bachmann2022multimae}, and recent driving-oriented approaches such as ALSO, SLidR, UniPAD, and GeoMAE~\cite{boulch2023also,sautier2022slidr,yang2024unipad,tian2023geomae}.
Compared with reconstruction-based and pairwise contrastive objectives, \method uses SIGReg's isotropic Gaussian target as a shared regularizing anchor while learnable fusion tokens provide a structured mechanism for cross-modal interaction inside the transformer.

\section{Method}
\label{sec:method}

We present \method, a multi-modal self-supervised framework that extends LeJEPA~\cite{balestriero2025lejepa} to jointly learn from RGB images and aligned companion modalities.
\Cref{fig:architecture} provides an overview of the architecture.

\subsection{Preliminaries: LeJEPA and SIGReg}
\label{sec:lejepa_prelim}

LeJEPA~\cite{balestriero2025lejepa} is a Joint-Embedding Predictive Architecture that learns invariance across augmented views while preventing representation collapse through \emph{Sketched Isotropic Gaussian Regularization} (SIGReg).

Let $\{\mathbf{x}_v^{g}\}_{v=1}^{V_g}$ denote the $V_g$ global augmented image views of an input and $\{\mathbf{x}_u^{\ell}\}_{u=1}^{V_\ell}$ its $V_\ell$ local views.
Each view is first processed by the shared encoder $f_\theta$.
The resulting representation is then mapped by the projector $g_\phi$ to latent embeddings $\mathbf{z}_v^{g} = g_\phi(f_\theta(\mathbf{x}_v^{g}))$ and $\mathbf{z}_u^{\ell} = g_\phi(f_\theta(\mathbf{x}_u^{\ell}))$ in $\mathbb{R}^d$.
The training objective combines an invariance loss with SIGReg:
\begin{equation}
    \mathcal{L}_\text{LeJEPA} = \lambda \cdot \mathcal{L}_\text{SIGReg}(\mathbf{Z}) + (1 - \lambda) \cdot \mathcal{L}_\text{inv},
    \label{eq:lejepa}
\end{equation}
where $\lambda$ is a single trade-off hyperparameter.
In the multimodal implementation used for \method, the target center is computed from the global views and the penalty is applied to all available views, so both global and local crops are pulled toward a shared center.
SIGReg then prevents collapse by matching the empirical embedding distribution to $\mathcal{N}(0, \mathbf{I})$ through characteristic-function matching over random projections and fixed evaluation knots.
This yields a regularizer with $\mathcal{O}(BK(T+d))$ complexity, where $B$ is the batch size, $K$ the number of random projection directions, $T$ the number of evaluation knots, and $d$ the embedding dimension, with no stop-gradients or teacher-student networks required; the detailed invariance and SIGReg expressions are given in the supplementary material.

\subsection{Multi-Modal Inputs}
\label{sec:input_repr}

To enable a unified transformer-based encoder for both modalities, we represent camera images and aligned companion signals in a common 2D format.

\paragraph{Camera Input.}
RGB images are processed using standard ViT patch embedding~\cite{dosovitskiy2021vit}.
Each image is resized to $224 \times 224$ and divided into $N = 14 \times 14 = 196$ non-overlapping patches of size $16 \times 16$, which are linearly projected to the ViT embedding dimension.
We apply the multi-crop augmentation strategy from LeJEPA~\cite{balestriero2025lejepa}: $V$ global crops (scale $[0.4, 1.0]$, size $224 \times 224$) and $V_{\text{local}}$ local crops (scale $[0.05, 0.4]$, size $96 \times 96$), together with ColorJitter, RandomGrayscale, GaussianBlur, and RandomSolarize augmentations.

\paragraph{Companion-Modality Input.}
For RGB-depth experiments, we project the 3D LiDAR point cloud into the camera coordinate frame to obtain an \emph{aligned depth map}.
We render that depth map by writing points in depth-descending order so nearer points overwrite farther ones, normalized by a maximum range of $r_\text{max} = 80$\,m; pixels with no LiDAR return remain zero.
This discards some native 3D structure, but it gives strict pixel-level alignment with RGB, keeps both modalities inside a unified transformer tokenization scheme, and avoids a separate 3D backbone.
For RGB-thermal experiments, the companion signal is the aligned thermal image itself, resized and patchified on the same spatial grid through its own modality-specific patch stem.
Using a shared 2D tokenization lets the same dense ViT family cover both RGB-depth and RGB-thermal settings rather than mixing image-plane transformers with a separate sparse-3D architecture only for LiDAR.

\paragraph{Modality Embeddings.}
To distinguish between camera and companion-modality tokens in the shared transformer, we add learnable modality embeddings $\mathbf{e}_\text{cam}$ and $\mathbf{e}_\text{mod}$ to the respective patch embeddings before entering the transformer blocks.
After tokenization, these become the camera token set $\mathbf{C}$ and companion-modality token set $\mathbf{M}$ used in \cref{eq:token_layout}.

\subsection{Learnable Fusion Tokens}
\label{sec:fusion_tokens}

A central design choice in multi-modal transformers is how and where to fuse information across modalities.
Rather than relying on full-token concatenation or late fusion, we follow the logic of a latent bottleneck~\cite{jaegle2021perceiverio}: a learned token set mediates cross-modal exchange, replacing expensive camera-to-LiDAR all-pairs interaction with routing through compact fusion tokens that act as a spatial memory buffer.

We introduce \emph{learnable fusion tokens} that provide a structured mechanism for cross-modal interaction.
Given $N$ spatial patch positions, we create $N$ learnable fusion token embeddings $\{\mathbf{f}_i\}_{i=1}^{N}$, each associated with a spatial position and represented in the same embedding space $\mathbb{R}^{D}$ as the ViT tokens.
We keep the number of fusion tokens equal to the patch count because the current design preserves one latent per spatial location: this keeps the cross-modal pairing explicit in the first fusion layer and retains the full $14\times14 = 196$ spatial grid that can be read out directly by the downstream patch probes.
For the from-scratch experiments in this paper, these fusion tokens are initialized with a truncated normal distribution with standard deviation $0.02$, matching the ViT token initialization used elsewhere in the encoder.
Together with a shared CLS token and the $N$ camera and $N$ companion-modality patch tokens, the full token sequence entering the transformer is:
\begin{equation}
    [\texttt{CLS}(1),\; \mathbf{F}(N),\; \mathbf{C}(N),\; \mathbf{M}(N)],
    \label{eq:token_layout}
\end{equation}
where $\mathbf{F}$, $\mathbf{C}$, $\mathbf{M}$ denote fusion, RGB, and companion-modality token sets respectively, for a total of $1 + 3N$ tokens (589 tokens for the $14\times14$ patch grid used here).
We extend the positional embedding to cover all tokens.

\paragraph{Attention Masking Strategies.}
We consider the default pruned design together with a persistent-routing ablation that controls how fusion tokens interact with modality tokens across layers.

\noindent\textbf{(a) Pruned Fusion.} In the first transformer layer (layer 0), fusion token $\mathbf{f}_i$ is allowed to attend to its spatially corresponding camera patch $\mathbf{c}_i$ and companion-modality patch $\mathbf{m}_i$, as well as to itself and the CLS token.
After layer 0, all $2N$ camera and companion-modality tokens are pruned from the sequence, leaving only $1 + N$ tokens (CLS + fusion).
Because the fusion tokens attend to the modality-specific patches in layer 0, gradients still flow back through that cross-attention path to update both the RGB and companion-modality patch stems even though those tokens are pruned later.
This creates an efficient latent bottleneck and reduces the computational cost of subsequent layers from $\mathcal{O}((1+3N)^2)$ to $\mathcal{O}((1+N)^2)$ per layer, a $\sim$9$\times$ reduction in attention cost, while forcing the model to compress the useful cross-modal evidence into the shared fusion-token grid early.

\noindent\textbf{(b) Persistent Fusion (ablation).} Fusion tokens maintain attention to their paired camera and companion-modality patches throughout \emph{all} transformer layers.
This preserves the full token set and allows deeper cross-modal reasoning at higher computational cost.
Unless noted otherwise, \method refers to the default FT-Pruned SIGReg configuration, i.e., the pruned fusion-token encoder trained with the joint-CLS objective below.

\subsection{Training Objective}
\label{sec:training_obj}

The default \method training objective keeps the encoder and token routing identical to \cref{sec:fusion_tokens} but applies JEPA supervision only to the joint multimodal CLS embedding.
For each paired input, the encoder processes the full multimodal token sequence once and produces a single fused CLS embedding that is passed through the projection head.
The spatial fusion tokens are \emph{not} projected individually. Instead, they are updated indirectly because the CLS token aggregates information from the fusion-token grid inside the same transformer, so the SIGReg and invariance gradients backpropagate through the CLS-to-fusion attention pathway. These fusion tokens are exactly the dense spatial features exposed to the downstream patch probes.
Let $\mathbf{Z}^{(\text{joint})}$ denote the resulting set of projected views from the multimodal crops.
The default loss is then
\begin{equation}
    \mathcal{L}_\text{MM} = \lambda \cdot \mathcal{L}_\text{SIGReg}(\mathbf{Z}^{(\text{joint})}) + (1 - \lambda) \cdot \mathcal{L}_\text{inv}^{(\text{joint})}.
    \label{eq:mm_loss}
\end{equation}
Here, $\mathcal{L}_\text{inv}^{(\text{joint})}$ is the mean-squared invariance term that pulls the projected global and local fused-crop embeddings toward the shared global-view center.
The computationally heavier FT-Pruned + SIGReg (3-pass) ablation keeps the same pruned encoder but averages SIGReg over three forwards: joint RGB+companion-modality, RGB-only with the companion modality zeroed out, and companion-modality-only with RGB zeroed out.
Alternative supervision variants are evaluated in the ablation study in \cref{tab:fusion_focus,tab:compute}, with the full 3-pass expression given in the supplementary material.

\section{Experiments}
\label{sec:experiments}

\subsection{Experimental Setup}
\label{sec:setup}

\begin{figure}[t]
\centering
\includegraphics[width=\linewidth]{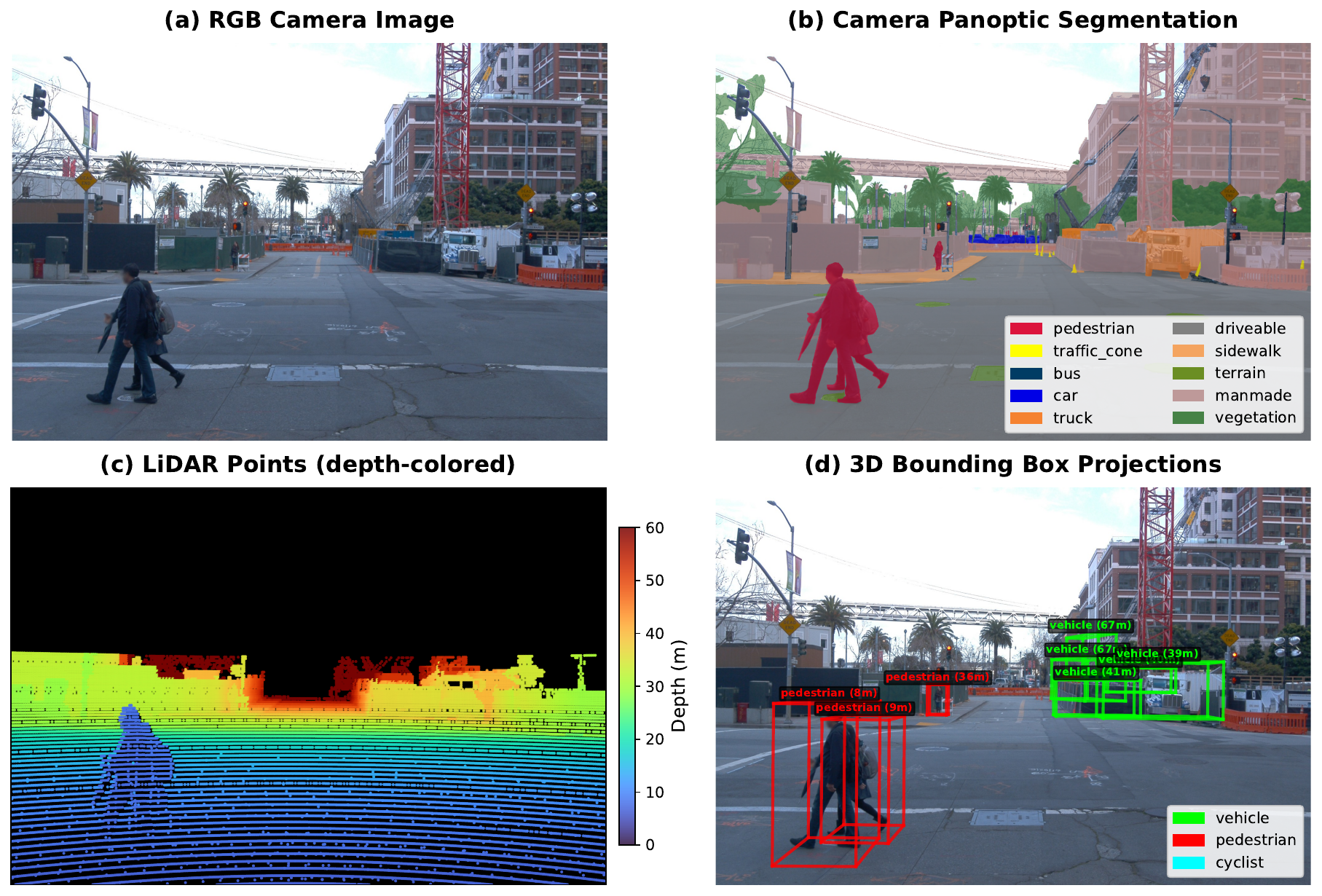}
\caption{\textbf{Waymo dataset showcase.} Example synchronized supervision used in our experiments: (a) RGB image, (b) camera-view segmentation, (c) the aligned companion-modality signal shown in depth form for the driving setting, and (d) projected 3D bounding boxes.}
\label{fig:waymo_dataset_showcase}
\end{figure}

\paragraph{Dataset.}
We use Waymo~\cite{sun2020waymo} as the main driving benchmark, nuScenes~\cite{caesar2020nuscenes} as a second from-scratch driving benchmark, and the Teledyne FLIR ADAS dataset~\cite{flir_adas_v2} for RGB-thermal experiments that include both from-scratch training and Waymo$\rightarrow$FLIR transfer/fine-tuning.
Following the data preparation used for the Waymo experiments, we keep the full segment set but subsample the synchronized stream to 2 Hz, with camera-view supervision attached wherever labels are available for the retained frames and probe caches.
For nuScenes, the camera-view segmentation targets are generated by projecting the official lidarseg point labels into the image plane, so these masks are inherently sparser and noisier than the denser Waymo camera-view labels.
For the patch-based 3D detection probes, Waymo annotations are remapped to the three foreground classes used by our probe export: car, pedestrian, and cyclist.
Across datasets, the reported supervision covers 3D boxes, depth, and segmentation on Waymo; 3D boxes, depth, and projected segmentation on nuScenes; and 2D detection on FLIR, which does not provide segmentation labels.
\Cref{fig:waymo_dataset_showcase} shows the synchronized RGB, dense companion-modality, and projected-box views used throughout the Waymo experiments.

\paragraph{Implementation Details.}
Unless noted otherwise, all main experiments use a ViT-Small/16 backbone~\cite{dosovitskiy2021vit} implemented with \texttt{timm}~\cite{wightman2019timm} and batch size 64.
Waymo and nuScenes scratch runs use 5 SSL epochs followed by 5 probe-training epochs.
During self-supervised pretraining, the encoder sees $224\times224$ global crops and $96\times96$ local crops; frozen-probe evaluation instead uses deterministic clean probe views at $640\times640$.
For from-scratch FLIR, we use a longer 20-epoch schedule; the Waymo$\rightarrow$FLIR frozen-transfer block keeps the pretrained encoder fixed, and the end-to-end FLIR fine-tuning block uses the separate schedule summarized in the supplementary material.
In practice, both the SSL loss and the frozen-probe validation curves largely stabilized within this budget, so we keep the shorter schedule rather than stretching all runs to much longer image-only SSL recipes.
The default \method model is the pruned fusion-token encoder trained with SIGReg on the joint multimodal CLS embedding.
Baseline-specific settings are deferred to the supplementary material.

\paragraph{Patch Probes.}
The main paper reports only patch-level probes.
These probes are trained on frozen patch embeddings and include a CenterNet-style 3D detection head~\cite{zhou2019centernet}, dense depth prediction, semantic segmentation, and 2D detection heads.
They should therefore be read as representation-quality tests of the learned spatial embedding geometry rather than as end-to-end finetuned perception-system numbers.
In the frozen-probe benchmarks, every SSL encoder is evaluated with the same probe architecture and optimization protocol, which isolates encoder quality rather than detector-specific tuning; the FLIR fine-tuning rows later in this section provide a separate practical check that the same pretrained encoder also improves end-to-end downstream performance.
For dual-stream multimodal baselines, the reported benchmarks use the camera-aligned RGB patch readout so that the downstream probe interface and dimensionality remain fixed across methods.
In pilot ablations, simple post-hoc patch fusion strategies such as concatenation, averaging, and learned projection severely destabilized segmentation transfer, while CenterNet and high-resolution depth results changed only marginally.
We therefore report CenterNet-style 3D detection together with Depth MAE and Seg. mIoU on Waymo, the same patch-level metrics on nuScenes, and 2D detection on FLIR.
Detailed probe implementation notes are provided in the supplementary material.

\paragraph{Metric Definitions.}
All main detection numbers come from the CenterNet-style probe.
We report XY-match mAP and XZ-match mAP on Waymo and nuScenes, corresponding to Bird's-Eye-View XY-plane and elevation XZ-plane matching respectively, together with Depth MAE and Seg. mIoU where available, and mAP50 and Car mAP50 on FLIR; the exact metric definitions are given in the supplementary material.

\paragraph{Baselines and Run Selection.}
For the Waymo and nuScenes patch-probe comparisons, all encoders in the main tables are trained from scratch within the target dataset and evaluated under the same frozen-probe protocol.
We compare against LeJEPA and LiDAR-only JEPA controls, a DINOv3-style RGB baseline~\cite{simeoni2025dinov3}, MultiMAE and ImageBind multimodal baselines~\cite{bachmann2022multimae,girdhar2023imagebind}, and simple early- and late-fusion ablations implemented in the same training pipeline.
We exclude methods that rely on native 3D sparse-convolution or point-cloud backbones~\cite{boulch2023also,sautier2022slidr,yang2024unipad,tian2023geomae} so the comparison remains apples-to-apples among unified 2D transformer encoders.
For dual-stream multimodal baselines, this protocol keeps the downstream probe architecture fixed by using a camera-aligned patch readout rather than introducing a method-specific post-hoc fusion block only at evaluation time.
Foundation-model baselines such as ImageBind and MultiMAE are retrained strictly from scratch for the corresponding benchmark rather than evaluated as off-the-shelf pretrained models. FLIR is the only section that additionally reports Waymo-initialized transfer and fine-tuning. ImageBind serves as the RGB-depth dual-encoder contrastive baseline, while MultiMAE appears as self-supervised and multitask reconstruction variants. This keeps the comparison tied to objective families under identical data and compute constraints.

\subsection{Patch-Level Results}
\label{sec:patch_results}

\begin{table}[t]
\centering
\caption{\textbf{Self-supervised single-modality encoders vs. encoder-frozen controls on Waymo.} All rows train the probes. Rows labeled ``encoder frozen'' keep a randomly initialized encoder fixed, whereas the other rows train the encoder self-supervised before frozen-probe evaluation. We report XY mAP, Depth MAE, and Seg. mIoU.}
\label{tab:frozen_compare}
\vspace{0.2em}
\scriptsize
\setlength{\tabcolsep}{3pt}
\begin{tabular}{@{}p{0.48\columnwidth} c c c@{}}
\toprule
\textbf{Method} & \textbf{mAP XY} $\uparrow$ & \textbf{Depth MAE} $\downarrow$ & \textbf{Seg. mIoU} $\uparrow$ \\
\midrule
LeJEPA~\cite{balestriero2025lejepa}                & \textbf{19.3} & \underline{4.704} & \textbf{0.261} \\
LeJEPA encoder frozen~\cite{balestriero2025lejepa} & 13.8 & 6.734 & 0.126 \\
LiDAR-only                                          & \underline{15.4} & \textbf{2.982} & \underline{0.151} \\
LiDAR-only encoder frozen                           & 8.0 & 5.397 & 0.089 \\
\bottomrule
\end{tabular}
\end{table}

\begin{table}[t]
\centering
\caption{\textbf{Focused fusion comparison on Waymo for the fusion variants.} We compare the default \method setting against early and late fusion, persistent routing, VICReg, and the 3-pass SIGReg variant.}
\label{tab:fusion_focus}
\vspace{0.2em}
\scriptsize
\setlength{\tabcolsep}{3pt}
\begin{tabular}{@{}p{0.52\columnwidth} c c c@{}}
\toprule
\textbf{Method} & \textbf{mAP XY} $\uparrow$ & \textbf{Depth MAE} $\downarrow$ & \textbf{Seg. mIoU} $\uparrow$ \\
\midrule
Early Fusion RGBD                           & 18.1 & 4.767 & 0.248 \\
Late Fusion                                 & 18.7 & 4.802 & 0.251 \\
FT-Pruned + VICReg~\cite{bardes2022vicreg} & 22.8 & 2.911 & 0.248 \\
FT-Persistent + SIGReg                      & 23.1 & \underline{2.846} & 0.271 \\
FT-Pruned + SIGReg (3-pass)                 & \underline{23.2} & \textbf{2.777} & \underline{0.274} \\
\method                                     & \textbf{23.6} & 2.860 & \textbf{0.275} \\
\bottomrule
\end{tabular}
\end{table}

\begin{table*}[t]
\centering
\caption{\textbf{Patch-probe comparison on Waymo with all encoders trained from scratch.} CenterNet and dense prediction heads are trained on frozen patch embeddings taken from models that were all self-supervised from scratch on Waymo. ``Training Data'' denotes the modalities seen during SSL pretraining; for dual-stream baselines, the frozen patch probes use the camera-aligned RGB patch readout described in \Cref{sec:setup}. Best results are in \textbf{bold}; second best are \underline{underlined}. $\uparrow$ means higher is better and $\downarrow$ means lower is better.}
\label{tab:main_results}
\vspace{0.5em}
\small
\setlength{\tabcolsep}{5pt}
\begin{tabular}{@{}l c c c c c@{}}
\toprule
\textbf{Method} & \textbf{Training Data} & \textbf{mAP XY} $\uparrow$ & \textbf{Depth MAE} $\downarrow$ & \textbf{mAP XZ} $\uparrow$ & \textbf{Seg. mIoU} $\uparrow$ \\
\midrule
LeJEPA~\cite{balestriero2025lejepa}      & RGB         & \underline{19.3} & 4.704 & \underline{4.9} & 0.261 \\
DINOv3~\cite{simeoni2025dinov3}          & RGB         & 15.2 & 5.314 & 3.5 & 0.239 \\
LiDAR-only                                & Depth       & 15.4 & \underline{2.982} & 4.0 & 0.151 \\
MultiMAE-SS~\cite{bachmann2022multimae}  & RGB+Depth   & 13.5 & 4.441 & 2.7 & 0.221 \\
MultiMAE-MT~\cite{bachmann2022multimae}  & RGB+Depth   & 13.7 & 3.583 & 2.9 & \underline{0.262} \\
ImageBind~\cite{girdhar2023imagebind}    & RGB+Depth   & 13.4 & 4.309 & 3.0 & 0.243 \\
\method                                   & RGB+Depth   & \textbf{23.6} & \textbf{2.860} & \textbf{7.2} & \textbf{0.275} \\
\bottomrule
\end{tabular}
\end{table*}

\Cref{tab:frozen_compare} compares self-supervised single-modality encoders against their randomly initialized, encoder-frozen counterparts.
Allowing the encoder to train under the LeJEPA objective substantially improves both modalities over keeping the encoder frozen at random initialization, which shows the value of learning the representation rather than relying only on the downstream probe.
The difference between RGB and LiDAR is also expected: RGB is stronger on localization and segmentation cues, whereas LiDAR is naturally stronger on depth reconstruction.

\Cref{tab:fusion_focus} then isolates the architectural comparison.
All three token-based fusion variants are much better than early or late fusion on depth.
The key takeaway from this table is that the default \method configuration is the strongest overall row: it gives the best XY mAP and Seg. mIoU while staying close to the best depth row and ahead of the persistent variant on the overall accuracy-efficiency balance.
We attribute this to an explicit information bottleneck: because the modality-specific tokens disappear after the first layer, the model is forced to compress the useful cross-modal evidence into the shared fusion-token grid early, whereas persistent routing can spend capacity on repeatedly revisiting redundant paired tokens.
The smaller but consistent gap between FT-Pruned + VICReg and the default SIGReg row suggests that the isotropic Gaussian target is also a better regularizing anchor for the joint multimodal CLS embedding than variance-and-covariance matching alone, likely because it more directly discourages modality-specific anisotropy in the shared latent space.

\Cref{tab:main_results} broadens the comparison to the full baseline set.
This is the paper's primary patch-probe benchmark.
Because \Cref{tab:main_results} compares encoders that are all trained from scratch on the same Waymo setup, the ranking is easier to interpret: \method achieves the best performance across all four spatial metrics in this comparison.
These absolute detection values are intentionally conservative because they come from frozen patch probes on top of fixed self-supervised features rather than from end-to-end detector fine-tuning.
The takeaway from this table is that the multimodal JEPA objective yields the strongest overall object-centric representation in the comparison.

MultiMAE-MT remains the strongest reconstruction-style reference on Waymo, which is consistent with the extra segmentation supervision available in that pretraining setup, while ImageBind remains a useful contrastive multimodal control but does not lead any of the main Waymo columns. That the multimodal foundation baselines do not clearly surpass the RGB-only LeJEPA control likely reflects the higher data demands of contrastive and masked-reconstruction objectives under strictly from-scratch training on our Waymo subset.
LiDAR-only and the DINOv3-style RGB baseline both remain clearly below \method, indicating that neither single-modality geometry nor stronger RGB-only pretraining is sufficient to match joint representation learning.

\subsection{nuScenes Results}
\label{sec:nuscenes_results}

\begin{table}[t]
\centering
\caption{\textbf{Results on the nuScenes dataset.} We report CenterNet XY-match mAP, XZ-match mAP, semantic segmentation, and Depth MAE.}
\label{tab:nuscenes}
\vspace{0.2em}
\scriptsize
\setlength{\tabcolsep}{2pt}
\begin{tabular}{@{}p{0.28\columnwidth} c c c c@{}}
\toprule
\textbf{Method} & \textbf{mAP XY} $\uparrow$ & \textbf{mAP XZ} $\uparrow$ & \textbf{Seg. mIoU} $\uparrow$ & \textbf{Depth MAE} $\downarrow$ \\
\midrule
MultiMAE-SS~\cite{bachmann2022multimae} & \underline{6.95} & \underline{1.66} & 0.195 & 5.736 \\
MultiMAE-MT~\cite{bachmann2022multimae} & 6.67 & 1.62 & 0.192 & \underline{5.624} \\
ImageBind~\cite{girdhar2023imagebind}   & 6.86 & 1.57 & \underline{0.198} & 5.912 \\
\method                                  & \textbf{9.52} & \textbf{2.53} & \textbf{0.228} & \textbf{2.031} \\
\bottomrule
\end{tabular}
\end{table}

\Cref{tab:nuscenes} shows that the default model remains the strongest row when trained from scratch on nuScenes.
The same fusion-token design that wins on Waymo also gives the best XY-match mAP, the best XZ-match mAP, the strongest segmentation score, and by far the best depth error on nuScenes.
The extra segmentation supervision used by MultiMAE-MT is less explicit here than on Waymo, which is unsurprising given that the nuScenes camera-view segmentation targets are projected from lidarseg labels and therefore noisier than the denser Waymo masks.

\subsection{FLIR Results}
\label{sec:flir_results}

\begin{table}[t]
\centering
\caption{\textbf{FLIR results across training regimes.} We report 2D CenterNet mAP50 and Car mAP50. Rows are ordered from training from scratch on FLIR, to Waymo$\rightarrow$FLIR transfer, to Waymo-pretrained fine-tuning on FLIR. Because FLIR does not provide segmentation labels, the scratch block excludes MultiMAE-MT.}
\label{tab:flir}
\vspace{0.2em}
\scriptsize
\setlength{\tabcolsep}{4pt}
\begin{tabular}{@{}l l c c@{}}
\toprule
\textbf{Setting} & \textbf{Method} & \textbf{mAP50} $\uparrow$ & \textbf{Car mAP50} $\uparrow$ \\
\midrule
Scratch & MultiMAE-SS~\cite{bachmann2022multimae} & \textbf{0.51} & \textbf{3.86} \\
Scratch & ImageBind~\cite{girdhar2023imagebind}   & 0.24 & 1.90 \\
Scratch & \method                                  & \underline{0.41} & \underline{3.00} \\
\midrule
Waymo$\rightarrow$FLIR & MultiMAE-SS~\cite{bachmann2022multimae} & 0.54 & 4.21 \\
Waymo$\rightarrow$FLIR & MultiMAE-MT~\cite{bachmann2022multimae} & 0.69 & 5.26 \\
Waymo$\rightarrow$FLIR & ImageBind~\cite{girdhar2023imagebind}   & \underline{0.72} & \underline{5.49} \\
Waymo$\rightarrow$FLIR & \method                                  & \textbf{1.56} & \textbf{10.22} \\
\midrule
Fine-tune & MultiMAE-SS~\cite{bachmann2022multimae} & 0.63 & 4.81 \\
Fine-tune & MultiMAE-MT~\cite{bachmann2022multimae} & \underline{0.75} & \underline{5.83} \\
Fine-tune & ImageBind~\cite{girdhar2023imagebind}   & 0.71 & 5.54 \\
Fine-tune & \method                                  & \textbf{2.39} & \textbf{12.88} \\
\bottomrule
\end{tabular}
\end{table}

\Cref{tab:flir} exposes two useful patterns.
First, the Waymo-pretrained \method encoder transfers more effectively to FLIR than the other multimodal baselines without any FLIR fine-tuning.
Second, the gap widens further after fine-tuning: \method reaches the best 2D detection transfer by a clear margin, while training from scratch on FLIR remains weak and noisy for all methods, likely because FLIR is much smaller than Waymo.

\subsection{Fusion and Loss Ablations Through Compute Profiling}
\label{sec:compute_ablation}

\begin{table}[t]
\centering
\caption{\textbf{Encoder-side compute profile.} Rows are sorted by estimated total encoder training time. We report time in minutes, total encoder SSL FLOPs in PFLOPs, and peak reserved VRAM in GB. Time is estimated for the encoder-side workload on an H200 system with an AMD EPYC 9275F 24-Core Processor. The fusion-token rows use the default single-pass joint-CLS objective, with the 3-pass SIGReg variant shown separately as an ablation.}
\label{tab:compute}
\vspace{0.3em}
\scriptsize
\setlength{\tabcolsep}{3.5pt}
\resizebox{\columnwidth}{!}{%
\begin{tabular}{@{}l c c c@{}}
\toprule
\textbf{Method} & \textbf{Time (min)} $\downarrow$ & \textbf{SSL FLOPs (PFLOPs)} $\downarrow$ & \textbf{VRAM (GB)} $\downarrow$ \\
\midrule
MultiMAE-SS~\cite{bachmann2022multimae}  & \textbf{14.07} & \textbf{21.4}  & \textbf{2.94} \\
Early Fusion RGBD                         & \underline{18.70} & 72.9  & 6.15 \\
LeJEPA~\cite{balestriero2025lejepa}      & 18.73 & 72.7  & \underline{6.01} \\
ImageBind~\cite{girdhar2023imagebind}    & 23.65 & 80.4  & 9.49 \\
Late Fusion                               & 34.72 & 144.9 & 11.41 \\
DINOv3~\cite{simeoni2025dinov3}          & 37.72 & 108.8 & 8.70 \\
\method                                   & 55.09 & 86.0  & 12.11 \\
FT-Pruned + VICReg                        & 59.41 & 84.8  & 12.03 \\
MultiMAE-MT~\cite{bachmann2022multimae}  & 63.49 & \underline{56.9}  & 7.45 \\
FT-Pruned + SIGReg (3-pass)               & 165.98 & 258.0 & 33.98 \\
FT-Persistent + SIGReg                    & 170.63 & 218.0 & 36.08 \\
\bottomrule
\end{tabular}
}
\end{table}

\Cref{tab:compute} makes the practical trade-off explicit.
Relative to the 3-pass row, the default model cuts the estimated training time by almost $3\times$ while also reducing peak VRAM and total FLOPs.
Persistent paired fusion remains the expensive edge of the family even under the same joint-CLS loss because it retains paired cross-modal token interactions throughout all transformer layers instead of pruning the modality-specific tokens after the first layer; that deeper routing increases both activations and attention-state memory, and its modest depth gain does not offset the extra cost.

The broader compute table also helps interpret the baselines.
Early and late fusion are much cheaper than token-based multimodal routing, but \Cref{tab:fusion_focus} shows that the cheaper designs leave clear detection and depth performance on the table. The default \method row remains the best-performing point inside the stronger token-fusion family before the much more expensive 3-pass and persistent variants.

\subsection{Qualitative Results}
\label{sec:qualitative}

\begin{figure}[t] 
	\centering
	\includegraphics[width=\linewidth]{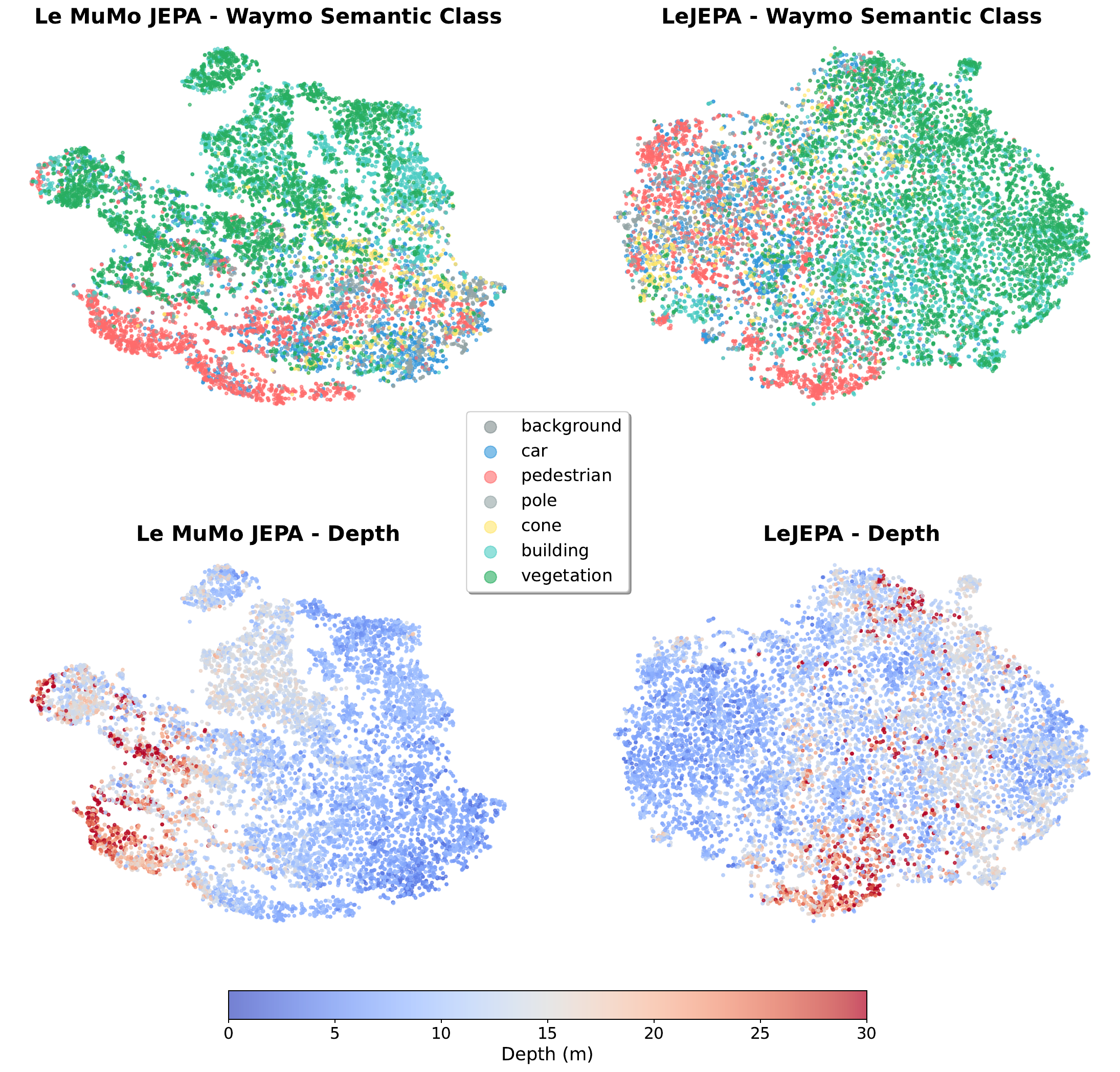}
	\caption{\textbf{Waymo patch-embedding visualization.} t-SNE projections of final-layer patch embeddings are shown with class-oriented structure and depth-oriented structure, illustrating how the learned patch space organizes both semantic grouping and geometric variation across methods. The left column uses \method fusion-token embeddings, and the right column uses LeJEPA patch embeddings. A planar fit to the plotted depth gradient gives an $R^2$ score of $0.463$ for \method versus $0.086$ for LeJEPA.}
	\label{fig:patch_comparison}
\end{figure}

\Cref{fig:patch_comparison} provides an embedding-space view that is consistent with the patch-probe tables: the fusion-token representation is more organized than the single-modality baseline when semantic grouping and geometry must be resolved jointly.
In the same 2D t-SNE space, a 10-NN class-purity score also improves from $0.453$ for the RGB-only LeJEPA baseline to $0.514$ for \method, against random same-class baselines of $0.215$ in both cases.
That depth-oriented structure is not only visual: fitting a simple planar trend in the 2D t-SNE space of the plotted depth gradient gives an $R^2$ score of $0.463$ for \method versus $0.086$ for the RGB-only LeJEPA baseline.

\begin{figure}[t]
	\centering
	\includegraphics[width=\linewidth]{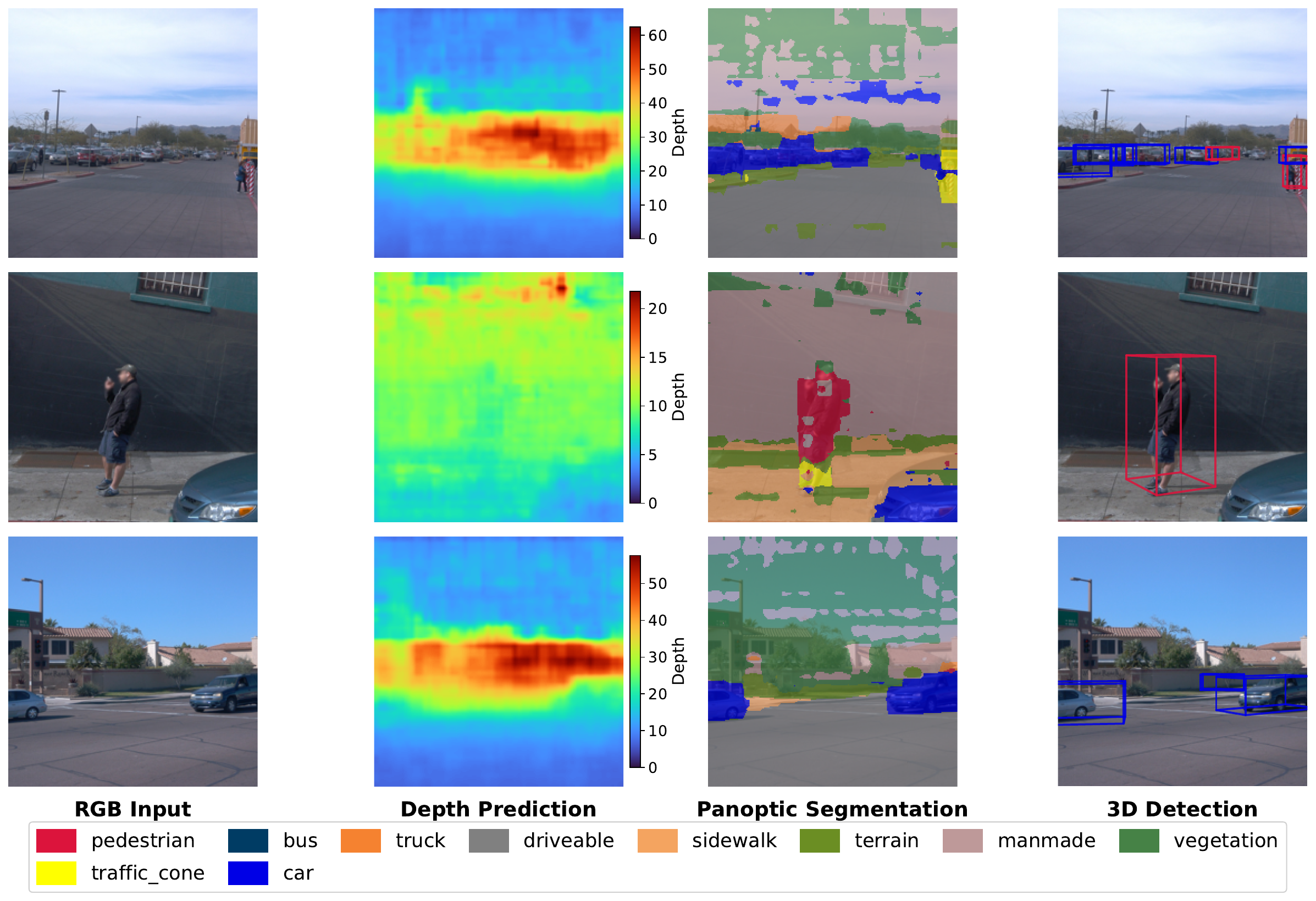}
	\caption{\textbf{Waymo qualitative probe output.} The figure shows the RGB input together with predictions from the three probe families used in the paper: dense depth estimation, segmentation, and 3D detection boxes. It provides a direct qualitative view of the same patch-level capabilities summarized by the quantitative probe tables.}
	\label{fig:probe_combined}
\end{figure}

\Cref{fig:probe_combined} complements the tables with direct Waymo prediction-level examples: the 3D boxes stay centered on vehicles, the depth prediction preserves the large layout changes between road and nearby objects, and the segmentation output maintains coherent road and sidewalk regions.

\section{Conclusion}
\label{sec:conclusion}

We presented \method, a multi-modal self-supervised framework that extends LeJEPA to jointly learn from RGB and aligned companion modalities.
By combining camera-aligned inputs, learnable fusion tokens, and SIGReg, \method enables structured multimodal representation learning in a shared token space without labeled pretraining data.

Our experiments show that the default \method configuration is now the strongest overall configuration in the paper.
On Waymo it leads the patch-probe benchmark, on nuScenes it also gives the best from-scratch row, and on FLIR it performs best under Waymo-initialized transfer and fine-tuning.
Among the fusion-token variants, the pruned default model also offers the best accuracy-efficiency balance while remaining far more practical than persistent routing in FLOPs, VRAM, and estimated training time.

More broadly, these results suggest that efficient shared-token multimodal pretraining can become a practical foundation for future autonomous driving systems that must integrate heterogeneous sensors without relying on expensive label-heavy supervision at every stage.

\subsection{Limitations and Broader Impacts}
Our evaluation is limited to frozen probes, a small dataset suite, and ViT-Small/16-scale backbones, and the current encoder relies on camera-aligned depth maps that simplify fusion but discard part of the native 3D structure.
The present formulation also benefits from reasonably aligned cross-modal observations, so robustness to calibration error, missing overlap, or weaker image-plane correspondence is not yet established.
Future work should pair the fusion-token design with sparse-voxel or point-cloud backbones, expand the nuScenes evaluation beyond detection, and test larger modality families beyond RGB-depth and RGB-thermal.
It should also verify how the learned fusion tokens scale to larger backbones.
Stronger multimodal pretraining could reduce annotation cost and improve robustness across sensing conditions, but deployment-facing claims would still require much broader safety validation and failure analysis than this probe-based study provides.

\section*{Acknowledgments}
This work received financial support from the Flanders AI Research Program (FAIR).
Code is available at \url{https://github.com/ciemcornelissen/le-mumo-jepa}.

{%
    \small
    \bibliographystyle{ieeenat_fullname}
    \bibliography{main}

@String(CVPR = {IEEE Conf. Comput. Vis. Pattern Recog.})

@String(ICCV = {Int. Conf. Comput. Vis.})

@String(ECCV = {Eur. Conf. Comput. Vis.})

@String(NIPS = {Adv. Neural Inform. Process. Syst.})

@String(ICLR = {Int. Conf. Learn. Represent.})

@article{balestriero2025lejepa,
  title     = {{LeJEPA}: Provable and Scalable Self-Supervised Learning Without the Heuristics},
  author    = {Balestriero, Randall and LeCun, Yann},
  journal   = {arXiv preprint arXiv:2511.08544},
  year      = {2025}
}

@inproceedings{assran2023ijepa,
  title     = {Self-Supervised Learning from Images with a Joint-Embedding Predictive Architecture},
  author    = {Assran, Mahmoud and Duval, Quentin and Misra, Ishan and Bojanowski, Piotr and Vincent, Pascal and Rabbat, Michael and LeCun, Yann and Ballas, Nicolas},
  booktitle = CVPR,
  pages     = {15619--15629},
  year      = {2023}
}

@article{bardes2024vjepa,
  title     = {Revisiting Feature Prediction for Learning Visual Representations from Video},
  author    = {Bardes, Adrien and Garrido, Quentin and Ponce, Jean and Chen, Xinlei and Rabbat, Michael and LeCun, Yann and Assran, Mahmoud and Ballas, Nicolas},
  journal   = {arXiv preprint arXiv:2404.08471},
  year      = {2024}
}

@inproceedings{chen2020simclr,
  title     = {A Simple Framework for Contrastive Learning of Visual Representations},
  author    = {Chen, Ting and Kornblith, Simon and Norouzi, Mohammad and Hinton, Geoffrey},
  booktitle = {Int. Conf. Mach. Learn.},
  pages     = {1597--1607},
  year      = {2020}
}

@inproceedings{he2020moco,
  title     = {Momentum Contrast for Unsupervised Visual Representation Learning},
  author    = {He, Kaiming and Fan, Haoqi and Wu, Yuxin and Xie, Saining and Girshick, Ross},
  booktitle = CVPR,
  pages     = {9729--9738},
  year      = {2020}
}

@inproceedings{grill2020byol,
  title     = {Bootstrap Your Own Latent - A New Approach to Self-Supervised Learning},
  author    = {Grill, Jean-Bastien and Strub, Florian and Altch\'{e}, Florent and Tallec, Corentin and Richemond, Pierre and Buchatskaya, Elena and Doersch, Carl and Avila Pires, Bernardo and Guo, Zhaohan and Gheshlaghi Azar, Mohammad and Piot, Bilal and Kavukcuoglu, Koray and Munos, Remi and Valko, Michal},
  booktitle = NIPS,
  editor    = {Larochelle, Hugo and Ranzato, Marc'Aurelio and Hadsell, Raia and Balcan, Maria-Florina and Lin, Hsuan-Tien},
  publisher = {Curran Associates, Inc.},
  pages     = {21271--21284},
  url       = {https://proceedings.neurips.cc/paper_files/paper/2020/file/f3ada80d5c4ee70142b17b8192b2958e-Paper.pdf},
  volume    = {33},
  year      = {2020}
}

@inproceedings{bardes2022vicreg,
  title     = {{VICReg}: Variance-Invariance-Covariance Regularization for Self-Supervised Learning},
  author    = {Bardes, Adrien and Ponce, Jean and LeCun, Yann},
  booktitle = ICLR,
  year      = {2022}
}

@inproceedings{caron2021dino,
  title     = {Emerging Properties in Self-Supervised Vision Transformers},
  author    = {Caron, Mathilde and Touvron, Hugo and Misra, Ishan and J\'{e}gou, Herv\'{e} and Mairal, Julien and Bojanowski, Piotr and Joulin, Armand},
  booktitle = ICCV,
  pages     = {9650--9660},
  year      = {2021}
}

@misc{simeoni2025dinov3,
  title         = {{DINOv3}},
  author        = {Sim\'{e}oni, Oriane and Vo, Huy V. and Seitzer, Maximilian and Baldassarre, Federico and Oquab, Maxime and Jose, Cijo and Khalidov, Vasil and Szafraniec, Marc and Yi, Seungeun and Ramamonjisoa, Micha\"{e}l and Massa, Francisco and Haziza, Daniel and Wehrstedt, Luca and Wang, Jianyuan and Darcet, Timoth\'{e}e and Moutakanni, Th\'{e}o and Sentana, Leonel and Roberts, Claire and Vedaldi, Andrea and Tolan, Jamie and Brandt, John and Couprie, Camille and Mairal, Julien and J\'{e}gou, Herv\'{e} and Labatut, Patrick and Bojanowski, Piotr},
  year          = {2025},
  eprint        = {2508.10104},
  archivePrefix = {arXiv},
  primaryClass  = {cs.CV},
  url           = {https://arxiv.org/abs/2508.10104}
}

@inproceedings{he2022mae,
  title     = {Masked Autoencoders Are Scalable Vision Learners},
  author    = {He, Kaiming and Chen, Xinlei and Xie, Saining and Li, Yanghao and Doll\'{a}r, Piotr and Girshick, Ross},
  booktitle = CVPR,
  pages     = {16000--16009},
  year      = {2022}
}

@inproceedings{bachmann2022multimae,
  title     = {{MultiMAE}: Multi-Modal Multi-Task Masked Autoencoders},
  author    = {Bachmann, Roman and Mizrahi, David and Atanov, Andrei and Zamir, Amir},
  booktitle = ECCV,
  pages     = {348--367},
  year      = {2022}
}

@inproceedings{bao2022beit,
  title     = {{BEiT}: {BERT} Pre-Training of Image Transformers},
  author    = {Bao, Hangbo and Dong, Li and Piao, Songhao and Wei, Furu},
  booktitle = ICLR,
  year      = {2022}
}

@inproceedings{dosovitskiy2021vit,
  title     = {An Image is Worth 16x16 Words: Transformers for Image Recognition at Scale},
  author    = {Dosovitskiy, Alexey and Beyer, Lucas and Kolesnikov, Alexander and Weissenborn, Dirk and Zhai, Xiaohua and Unterthiner, Thomas and Dehghani, Mostafa and Minderer, Matthias and Heigold, Georg and Gelly, Sylvain and Uszkoreit, Jakob and Houlsby, Neil},
  booktitle = ICLR,
  year      = {2021}
}

@inproceedings{jaegle2021perceiverio,
  title     = {Perceiver {IO}: A General Architecture for Structured Inputs \& Outputs},
  author    = {Jaegle, Andrew and Borgeaud, Sebastian and Alayrac, Jean-Baptiste and Doersch, Carl and Ionescu, Catalin and Ding, David and Koppula, Skanda and Zoran, Daniel and Brock, Andrew and Shelhamer, Evan and others},
  booktitle = ICLR,
  year      = {2022}
}

@article{wightman2019timm,
  title     = {{PyTorch} Image Models},
  author    = {Wightman, Ross},
  journal   = {GitHub repository},
  year      = {2019},
  note      = {\url{https://github.com/rwightman/pytorch-image-models}}
}

@inproceedings{liu2023bevfusion,
  title     = {{BEVFusion}: Multi-Task Multi-Sensor Fusion with Unified Bird's-Eye View Representation},
  author    = {Liu, Zhijian and Tang, Haotian and Amini, Alexander and Yang, Xinyu and Mao, Huizi and Rus, Daniela and Han, Song},
  booktitle = {Int. Conf. Robot. Autom.},
  pages     = {2774--2781},
  year      = {2023}
}

@article{liang2022bevfusion,
  title     = {{BEVFusion}: A Simple and Robust {LiDAR}-Camera Fusion Framework},
  author    = {Liang, Tingting and Xie, Hongwei and Yu, Kaicheng and Xia, Zhongyu and Lin, Zhiwei and Wang, Yongtao and Tang, Tao and Wang, Bing and Tang, Zhi},
  journal   = NIPS,
  volume    = {35},
  pages     = {10421--10434},
  year      = {2022}
}

@inproceedings{bai2022transfusion,
  title     = {{TransFusion}: Robust {LiDAR}-Camera Fusion for {3D} Object Detection with Transformers},
  author    = {Bai, Xuyang and Hu, Zeyu and Zhu, Xinge and Huang, Qingqiu and Chen, Yilun and Fu, Hongbo and Tai, Chiew-Lan},
  booktitle = CVPR,
  pages     = {1090--1099},
  year      = {2022}
}

@article{zhou2019centernet,
  title     = {Objects as Points},
  author    = {Zhou, Xingyi and Wang, Dequan and Kr\"{a}henb\"{u}hl, Philipp},
  journal   = {arXiv preprint arXiv:1904.07850},
  year      = {2019}
}

@inproceedings{sun2020waymo,
  title     = {Scalability in Perception for Autonomous Driving: {Waymo Open Dataset}},
  author    = {Sun, Pei and Kretzschmar, Henrik and Dotiwalla, Xerxes and Chouard, Aurelien and Patnaik, Vijaysai and Tsui, Paul and Guo, James and Zhou, Yin and Chai, Yuning and Caine, Benjamin and others},
  booktitle = CVPR,
  pages     = {2446--2454},
  year      = {2020}
}

@inproceedings{caesar2020nuscenes,
  title     = {{nuScenes}: A Multimodal Dataset for Autonomous Driving},
  author    = {Caesar, Holger and Bankiti, Varun and Lang, Alex H. and Vora, Sourabh and Liong, Venice Erin and Xu, Qiang and Krishnan, Anush and Pan, Yu and Baldan, Giancarlo and Beijbom, Oscar},
  booktitle = CVPR,
  pages     = {11621--11631},
  year      = {2020}
}

@misc{flir_adas_v2,
  author       = {{Teledyne FLIR LLC}},
  title        = {Teledyne FLIR Free ADAS Thermal Dataset v2},
  year         = {2022},
  howpublished = {\url{https://adas-dataset-v2.flirconservator.com/}},
  note         = {Accessed: 2026-03-19}
}

@inproceedings{girdhar2023imagebind,
  title     = {{ImageBind}: One Embedding Space To Bind Them All},
  author    = {Girdhar, Rohit and El-Nouby, Alaaeldin and Liu, Zhuang and Singh, Mannat and Alwala, Kalyan Vasudev and Joulin, Armand and Misra, Ishan},
  booktitle = CVPR,
  pages     = {15180--15190},
  year      = {2023}
}

@inproceedings{sautier2022slidr,
  title     = {Image-to-{LiDAR} Self-Supervised Distillation for Autonomous Driving Data},
  author    = {Sautier, Corentin and Puy, Gilles and Gidaris, Spyros and Boulch, Alexandre and Bursuc, Andrei and Marlet, Renaud},
  booktitle = CVPR,
  pages     = {9891--9901},
  year      = {2022}
}

@inproceedings{boulch2023also,
  title     = {{ALSO}: Automotive {LiDAR} Self-Supervision by Occupancy Estimation},
  author    = {Boulch, Alexandre and Sautier, Corentin and Michele, Bj\"{o}rn and Puy, Gilles and Marlet, Renaud},
  booktitle = CVPR,
  pages     = {13455--13465},
  year      = {2023}
}

@inproceedings{yang2024unipad,
  title     = {{UniPAD}: A Universal Pre-training Paradigm for Autonomous Driving},
  author    = {Yang, Honghui and Zhang, Sha and Huang, Di and Wu, Xiaoyang and Zhu, Haoyi and He, Tong and Tang, Shixiang and Zhao, Hengshuang and Qiu, Qibo and Lin, Binbin and He, Xiaofei and Ouyang, Wanli},
  booktitle = CVPR,
  pages     = {14746--14757},
  year      = {2024}
}

@inproceedings{tian2023geomae,
  title     = {{GeoMAE}: Masked Geometric Target Prediction for Self-Supervised Point Cloud Pre-Training},
  author    = {Tian, Xiaoyu and Ran, Haoxi and Wang, Yue and Zhao, Hang},
  booktitle = CVPR,
  pages     = {13570--13579},
  year      = {2023}
}

@inproceedings{vora2020pointpainting,
  title     = {{PointPainting}: Sequential Fusion for {3D} Object Detection},
  author    = {Vora, Sourabh and Lang, Alex H. and Helou, Bassam and Beijbom, Oscar},
  booktitle = CVPR,
  pages     = {4604--4612},
  year      = {2020}
}

@inproceedings{pang2020clocs,
  title     = {{CLOCs}: Camera-{LiDAR} Object Candidates Fusion for {3D} Object Detection},
  author    = {Pang, Su and Morris, Daniel and Radha, Hayder},
  booktitle = {IEEE/RSJ Int. Conf. Intell. Robot. Syst.},
  pages     = {10386--10393},
  year      = {2020}
}
}

\hypersetup{pageanchor=false}
\setcounter{section}{0}
\setcounter{subsection}{0}
\setcounter{table}{0}
\setcounter{figure}{0}
\setcounter{equation}{0}
\renewcommand{\thesection}{S\arabic{section}}
\renewcommand{\thesubsection}{S\arabic{section}.\arabic{subsection}}
\renewcommand{\thetable}{S\arabic{table}}
\renewcommand{\thefigure}{S\arabic{figure}}
\renewcommand{\theequation}{S\arabic{equation}}
\renewcommand{\theHsection}{supp.\arabic{section}}
\renewcommand{\theHsubsection}{supp.\arabic{section}.\arabic{subsection}}
\renewcommand{\theHtable}{supp.\arabic{table}}
\renewcommand{\theHfigure}{supp.\arabic{figure}}
\renewcommand{\theHequation}{supp.\arabic{equation}}
\def\continuesupplementpagecounter{1}
\clearpage
\ifdefined\continuesupplementpagecounter
\else
\setcounter{page}{1}
\fi
\maketitlesupplementary

\section{Baseline Implementation Details}
\label{sec:suppl_baselines}

This supplement records the implementation choices behind the main baselines used in \Cref{sec:experiments}.
The goal is not to reproduce every training flag inline in the main paper, but to make clear what each comparison represents and which prior work it is adapting.

\section{Detailed SIGReg Formulation}
\label{sec:suppl_sigreg}

In the multimodal implementation used for \method, the target center is computed from the global views and the penalty is applied to all available views:
\begin{equation}
	\begin{aligned}
	\bar{\mathbf{z}} &= \frac{1}{V_g}\sum_{v=1}^{V_g} \mathbf{z}_v^{g}, \\
	\mathcal{L}_\text{inv} &= \frac{1}{V_g + V_\ell} \left(
	\sum_{v=1}^{V_g} \|\mathbf{z}_v^{g} - \bar{\mathbf{z}}\|^2 +
	\sum_{u=1}^{V_\ell} \|\mathbf{z}_u^{\ell} - \bar{\mathbf{z}}\|^2
	\right).
	\end{aligned}
\end{equation}
SIGReg then matches the empirical embedding distribution to $\mathcal{N}(0, \mathbf{I})$ by projecting embeddings onto $K$ random directions and comparing the empirical characteristic function of those projections against the Gaussian target at $T$ evaluation knots:
\begin{equation}
	\begin{aligned}
	\hat{c}_{k,j} &= \frac{1}{B}\sum_{n=1}^{B}\cos\!\left(t_j\mathbf{w}_k^\top \mathbf{z}_n\right), \\
	\hat{s}_{k,j} &= \frac{1}{B}\sum_{n=1}^{B}\sin\!\left(t_j\mathbf{w}_k^\top \mathbf{z}_n\right), \\
	\mathcal{L}_\text{SIGReg}(\mathbf{Z}) &= \frac{1}{K}\sum_{k=1}^{K}\sum_{j=1}^{T} \omega_j
	\left[\left(\hat{c}_{k,j} - e^{-t_j^2/2}\right)^2 + \hat{s}_{k,j}^2\right].
	\end{aligned}
\end{equation}
These are the full expressions summarized more briefly in the main paper.

\paragraph{Shared training defaults.}
For the Waymo experiments used in the paper, the shared defaults are a ViT-Small backbone, batch size 64, 5 self-supervised training epochs, $V=2$ global crops, 8 local crops, projection dimension 16, learning rate $10^{-4}$, and $\lambda=0.1$ for SIGReg.
The self-supervised encoder is trained with $224\times224$ global crops and $96\times96$ local crops, whereas frozen-probe evaluation uses deterministic clean probe views at $640\times640$.
After encoder pretraining, the encoder is frozen and the probes are trained for 5 additional epochs, with validation every 100 steps on the full validation split.
Patch probes remain enabled during evaluation, and the occupancy IoU uses a neutral empty-union policy.
Unless noted otherwise, all baselines reuse the same data filtering, camera-view supervision, probe heads, validation cadence, and run-selection logic as the main method so that the comparison changes the representation learner rather than the downstream evaluation stack.

\paragraph{Modality-specific augmentations.}
The RGB stream receives the appearance-level augmentations listed in the main paper, namely ColorJitter, RandomGrayscale, GaussianBlur, and, in the official DINO-style branch, RandomSolarize.
The companion modality does \emph{not} receive those photometric perturbations.
For aligned RGB-depth training, the crop rectangle is sampled once and applied to both RGB and depth, and a single horizontal-flip decision is shared between the two streams; after that synchronized spatial step, RGB receives the photometric augmentation stack while the depth map only undergoes resizing/pooling, dtype conversion, and the shared flip.
For FLIR, the RGB and thermal crops are likewise sampled in a synchronized way and optionally flipped together, but the thermal branch uses only image conversion, float casting, and 1-channel normalization, while the RGB branch receives the JEPA/DINO-style photometric transforms.
This preserves pixel alignment across modalities without applying color-style perturbations to depth or thermal inputs, where they would not have a physical interpretation.

\paragraph{Waymo subset construction.}
The Waymo setup used throughout the paper is defined by the shared data-preparation pipeline rather than by manual scene curation.
For the reported runs, we keep the available segments for the selected split and subsample the synchronized stream from the native 10 Hz capture rate to 2 Hz, i.e., every fifth frame, exactly as described in the main paper.
The concrete export used by a run records the resulting segment and frame counts in the generated metadata, so the subset definition is deterministic even though those counts are not repeated inline in every table.

\paragraph{Tuning policy.}
We do not run a separate downstream hyperparameter search for each baseline.
The frozen-probe stage, deterministic probe views, validation cadence, run-selection rule, and probe heads are shared across methods, while baseline-specific changes are limited to objective-intrinsic settings such as DINO temperatures, InfoNCE temperature, or MultiMAE mask ratio and decoder size.
The goal is therefore to compare representation learners under the same short from-scratch budget rather than to maximize each baseline with method-specific probe engineering.

\paragraph{Single-modality JEPA baselines.}
\textbf{RGB-only} and \textbf{LiDAR-only} share the same basic encoder design, with the only architectural change being the input channel count: RGB uses a 3-channel input, whereas LiDAR depth uses a 1-channel input.
The paired \textbf{RGB-only frozen} and \textbf{LiDAR-only frozen} settings keep the encoder fixed and train only the probe heads.
These baselines therefore isolate cross-modal fusion from two different confounds at once: modality choice and encoder adaptation.
In particular, the trainable single-modality rows test whether the gains in the main paper come merely from stronger encoder optimization, whereas the frozen rows test how much downstream performance is available without any encoder-side adaptation at all.

\paragraph{Early and late fusion.}
\textbf{Early Fusion RGBD} uses a single encoder over stacked RGB and aligned depth channels.
\textbf{Late Fusion} uses separate modality encoders whose features are concatenated before probing.
These two settings are simple in-house architectural ablations within our training pipeline rather than direct reimplementations of specific prior supervised fusion systems.
Both baselines share the same Waymo data pipeline, probe heads, and compute profiling code as \method.
They are included to separate the benefit of multimodal data itself from the benefit of the learnable fusion-token bottleneck: early fusion tests whether naive channel stacking is sufficient, and late fusion tests whether keeping the modalities separate until readout is already enough.

\paragraph{DINOv3-style baseline.}
The DINO baseline used in the main table is a scratch-trained RGB model with a DINOv3-style training objective rather than a frozen pretrained encoder.
Its main hyperparameters are a DINO learning rate of $5\times 10^{-4}$, prototype dimension 1024, iBOT output dimension 1024, teacher temperature 0.04, teacher warmup start 0.04, zero DINO warmup epochs, and zero frozen-last-layer epochs.
The training setup additionally uses student temperature 0.1, teacher momentum 0.996, center momentum 0.9, and AdamW betas $(0.9, 0.95)$.
These are not intended to reproduce the official DINOv3 recipe exactly.
Instead, they are a tuned in-project configuration chosen to learn faster and remain competitive under the shorter from-scratch budget used throughout this paper.
This makes it a stronger RGB-only architectural baseline than plain JEPA, without introducing LiDAR or explicit multi-modal fusion.
Under the short from-scratch budget used in this paper, however, it still underperforms the simpler RGB-only LeJEPA control in the main table.
Its role in the paper is to test whether a stronger modern RGB-only SSL objective can close the gap to multimodal learning when both are trained under the same from-scratch protocol.

\paragraph{ImageBind-style baseline.}
The \textbf{ImageBind} baseline uses paired RGB-depth encoders trained with a symmetric InfoNCE objective at temperature 0.07.
This configuration disables local crops, uses clean probe views, runs at batch size 64 in the current rerun, and evaluates probes at high image resolution.
For the comparison reported in the main table, this baseline is trained in the same project pipeline as the other methods rather than being treated as a frozen pretrained encoder.
It therefore appears in both the main accuracy table and the compute table as a trainable multimodal baseline.
Conceptually, this row is the contrastive multimodal reference in the paper: two modality-specific encoders are aligned through paired-view InfoNCE rather than through predictive fusion tokens and JEPA-style prediction.
This is important for interpretation because it keeps the multimodal setting but changes the learning principle from predictive regularized representation learning to pairwise contrastive alignment.

\paragraph{MultiMAE baselines.}
\textbf{MultiMAE-SS} and \textbf{MultiMAE-MT} use a mask ratio of 0.75, decoder depth 2, and decoder width 256.
These variants disable local crops because the decoder expects a fixed global patch grid.
\textbf{MultiMAE-SS} is the self-supervised variant: it reconstructs multimodal RGB-depth content without using segmentation labels during representation learning.
\textbf{MultiMAE-MT} is the multitask variant: it keeps the same reconstruction backbone but additionally enables semantic supervision through the auxiliary labels that the Waymo pipeline already prepares.
These baselines are therefore intentionally stronger dense-prediction references than the simpler early- and late-fusion designs.
They serve as reconstruction-style multimodal baselines whose inductive bias is closer to masked modeling than to either contrastive alignment or JEPA-style latent prediction.

\paragraph{Fusion-token ablations.}
The compute table in the main paper includes \textbf{FT-Pruned}, \textbf{FT-Pruned + VICReg}, \textbf{FT-Persistent}, and \textbf{FT-Pruned + SIGReg (3-pass)} in addition to the default \textbf{\method} model, which corresponds to the FT-Pruned SIGReg setting used throughout the main comparison tables.
These scenarios all share the same fusion-token encoder family and differ mainly in their routing strategy and learning objective.
In the paper, the synchronized accuracy comparison focuses on the main pruned model, the VICReg variant, the persistent-routing variant, and the 3-pass objective, while the broader compute table shows the cost of different token-routing choices.
The ablations are intended to answer two separate questions: whether explicit token routing matters relative to simpler fusion, and whether the gains are specific to SIGReg on the joint embedding rather than to the encoder family alone.

\paragraph{Three-pass SIGReg ablation.}
The \textbf{FT-Pruned + SIGReg (3-pass)} row uses the same pruned fusion-token encoder as the default model, but it evaluates SIGReg on three forward passes instead of only the joint fused pass.
For each paired sample, it computes: (i) a joint RGB+companion-modality pass, (ii) an RGB-only pass with the companion modality zeroed out, and (iii) a companion-modality-only pass with RGB zeroed out.
If $\mathbf{Z}^{(\text{joint})}$, $\mathbf{Z}^{(\text{rgb})}$, and $\mathbf{Z}^{(\text{mod})}$ denote the projected CLS embeddings from those three passes, then the added three-pass regularizer is
\begin{equation}
	\begin{aligned}
	\mathcal{L}^{(3\text{-pass})}_{\text{SIGReg}}
	&= \frac{1}{3}\Big[
	\mathcal{L}_{\text{SIGReg}}(\mathbf{Z}^{(\text{joint})}) \\
	&\qquad + \mathcal{L}_{\text{SIGReg}}(\mathbf{Z}^{(\text{rgb})})
	+ \mathcal{L}_{\text{SIGReg}}(\mathbf{Z}^{(\text{mod})})
	\Big].
	\end{aligned}
\end{equation}
The joint pass is reused from the main objective, so the extra compute comes primarily from the two masked single-modality forwards.
This is the ablation referred to as ``3-pass'' in the Waymo and compute tables.

\paragraph{Compute-profile reporting.}
The compute table in the main paper is intentionally encoder-side only.
Its time column is the estimated encoder-training runtime from the shared logging pipeline on an H200 system with an AMD EPYC 9275F 24-Core Processor, and the FLOP and VRAM columns come from the same training logs via the profiled encoder SSL FLOPs and peak reserved GPU memory fields.
These numbers are meant to compare representation-learning cost under a common training stack; they do not include offline dataset preparation or the separate frozen-probe training stage.

\paragraph{Dataset-specific training details.}
\textbf{nuScenes from scratch.} These runs retrain the encoder directly on nuScenes for 5 SSL epochs and then train the frozen probes for 5 epochs with batch size 64, mirroring the short Waymo schedule under the same clean-view probe setup.

\textbf{Waymo$\rightarrow$FLIR frozen transfer.} These runs use the dedicated probe-evaluation configuration rather than the SSL pretraining loop: the pretrained Waymo encoder is frozen, probe-only training is enabled, $V=1$ and local crops are disabled, and only the downstream heads are optimized for 5 epochs with batch size 64.
The frozen-transfer block uses learning rate $10^{-4}$, probe learning rate $10^{-3}$, patch-probe learning rate $10^{-3}$, probe resolution $640\times640$, validation every 100 steps, and no LiDAR or copy-paste augmentation.

\textbf{FLIR from scratch.} The scratch FLIR rows use the longer 20-epoch FLIR schedule before the downstream evaluation stage, reflecting the smaller size of the RGB-thermal dataset.

\textbf{End-to-end FLIR fine-tuning.} The FLIR fine-tuning rows use the separate detection fine-tuning configuration with batch size 64, 30 epochs, encoder learning rate $2\times10^{-5}$, decoder learning rate $2\times10^{-4}$, AdamW with weight decay 0.05, validation every 100 steps, early-stopping patience 8, and random-resized-crop scale $[0.8, 1.0]$ in train mode.
The optimizer uses a 5\% linear warmup starting at 1\% of the target learning rate, followed by cosine decay to $10^{-7}$.
For FLIR, this configuration uses the 2D CenterNet-style detection path, a 3-layer decoder with hidden width 512, an auxiliary global-view size of 224, and a clean probe/evaluation view of $640\times640$.

\section{Probe Implementation Details}
\label{sec:suppl_probes}

This section focuses on the patch probes reported in the main paper.
All probes are trained after self-supervised pretraining with the encoder frozen, so the tables should be read as representation-quality measurements rather than as full end-to-end finetuning results.
The probe stage always uses the same deterministic $640\times640$ camera-view inputs so that differences in downstream numbers reflect the learned representation and not stochastic crop variation at evaluation time.

\paragraph{Patch probes.}
The patch-probe family contains a CenterNet-style 3D detection head with $2\times$ upsampling, segmentation heads, a dense depth-map probe with $4\times$ upsampling, and an occupancy-map probe.
All of these heads are shallow readouts rather than standalone perception backbones.
The stronger CenterNet-style 3D head keeps the spatial token grid, applies a $3\times3$ Conv$(\text{vit\_dim}, 256)$ adapter with batch normalization and ReLU, upsamples the $14\times14$ grid to $28\times28$ with a transposed convolution, and then uses separate $1\times1$ heads for heatmap, offset, size, depth, and yaw.
The segmentation probe is strictly linear: a single $1\times1$ convolution projects to $C\,r^2$ channels, PixelShuffle upsamples by $r=4$, and bilinear interpolation resizes the output to $224\times224$.
The depth-map probe is still lightweight but not purely linear: it uses a $1\times1$ projection to $16\,r^2$ channels, PixelShuffle with $r=4$, then a $3\times3$ refinement convolution and a final $1\times1$ depth head.
The occupancy-map probe uses a $1\times1$ projection branch, a $1\times1$ skip branch, PixelShuffle with $r=2$, two small $3\times3$ GroupNorm+GELU refinement blocks, and a final $1\times1$ prediction head.
Its loss is BCE-with-logits with focal reweighting, plus a Dice term and a small consistency term for multi-channel outputs.
For FLIR, the box-segmentation head reuses the same linear SemanticSegProbe template at occupancy-grid resolution, and the 2D detection head is the 2D analogue of the same CenterNet-style spatial readout.
For Waymo, the main paper reports the stronger CenterNet-style detector together with Depth MAE and Seg. mIoU.
The detector probes operate on frozen spatial features rather than on the global CLS token, which is why they are a better test of whether the learned representation preserves object layout, localization cues, and cross-modal geometry.

\paragraph{Patch-readout protocol for dual-stream baselines.}
For dual-stream multimodal baselines such as ImageBind and MultiMAE, the reported patch-probe benchmarks use a fixed camera-aligned patch readout rather than an additional post-hoc probe-time fusion module.
The reason is empirical rather than purely cosmetic: in pilot ablations, simple post-hoc multimodal patch fusion choices such as concatenation, averaging, and learned projection severely destabilized segmentation transfer, even when the higher-resolution depth probe and the CenterNet probe were less affected.
We therefore treat those probe-time fusion variants as unstable evaluation choices in the current frozen-feature setting and keep the main comparison focused on a matched probe interface rather than on method-specific readout engineering.

\paragraph{Metric definitions and correspondence.}
The main detection numbers come from the CenterNet-style spatial box probe.
For Waymo and nuScenes, the detection table columns are read from the CenterNet export as XY-match mAP and XY-match ADE, with XZ-match mAP additionally shown where that aggregate is available in the checked export.
These 3D detection mAP values are computed with the shared center-distance AP evaluation used throughout the codebase rather than with an official leaderboard submission script: for each evaluated class, AP is computed under center-distance matching at 0.5, 1, 2, and 4 meters, averaged over those four thresholds, and then averaged over classes.
In the Waymo patch-probe setting, the class average is over car, pedestrian, and cyclist; in the nuScenes from-scratch setting, it is over all evaluated classes in the export used for the table.
The main depth number is the higher-resolution dense depth-map MAE from the dedicated depth-map probe, reported as Depth MAE in the paper tables.
The main segmentation number is reported in the paper as Seg. mIoU. For Waymo, this value comes from the semantic-segmentation probe metric logged in the selected runs, while the nuScenes table uses the corresponding segmentation field from the export used for that table.
For FLIR, the main table switches to the available 2D detection outputs, namely CenterNet mAP50 and Car mAP50, because those are the directly comparable detection metrics logged for that dataset; in that setting, mAP50 is the mean AP at IoU 0.5 over all FLIR detection classes and Car mAP50 is the car-specific AP50.

\end{document}